\documentclass[10pt]{article} % For LaTeX2e
\usepackage[preprint]{tmlr}

\usepackage{natbib}
\usepackage{microtype}
\usepackage{graphicx}
\usepackage{subfigure}
\usepackage{booktabs}
\usepackage{hyperref}

\usepackage{amsmath}
\usepackage{amssymb}
\usepackage{mathtools}
\usepackage{amsthm}
\usepackage{scalerel}
\usepackage[shortlabels]{enumitem}

\usepackage[capitalize,noabbrev,nameinlink]{cleveref}
\creflabelformat{equation}{#2#1#3}
\usepackage{csquotes}
\usepackage{multirow}
\usepackage{mathrsfs}
\usepackage{mdframed}
\usepackage{bm}
\usepackage{bbm,tipa}
\usepackage{mathbbol}
\DeclareSymbolFontAlphabet{\mathbb}{AMSb}
\DeclareSymbolFontAlphabet{\mathbbl}{bbold}

\usepackage{xcolor}
\definecolor{dred}{RGB}{153,80,43}
\definecolor{dblue}{RGB}{0,114,178}
\hypersetup{
    colorlinks=true,
    breaklinks=true,
    urlcolor=dblue, %blue
    linkcolor=dred, %purple, violet, olive, brown
    citecolor=dblue
}
\usepackage{caption}

\newcommand{\mc}[1]{\mathcal{#1}}

\newcommand{\mb}[1]{\mathbbm{#1}}
\newcommand{\mr}[1]{\mathrm{#1}}

\newcommand{\ra}{\rightarrow}

\newcommand{\N}{\mb{N}}

\newcommand{\R}{\mb{R}}

\def\vmin{\mr{v}_{\mr{min}}}
\def\vmax{\mr{v}_{\mr{max}}}

\def\environment{e}
\def\env{\environment}
\def\environments{{\mc{E}}}

\def\actions{\mc{A}}
\def\observations{\mc{O}}
\def\states{\mc{S}}

\def\agent{\lambda}%f_{\lambda}}%
\def\agents{\Lambda}%\mc{F}_\lambda}%
\def\agentf{{\vec{\agent}}}
\def\allagents{\mathbbl{\agents}}
\newcommand\reaches{%
  \mathrel{\ooalign{\hss$\rightsquigarrow$\hss\kern-1.45ex\raise1.0ex\hbox{{\scaleobj{0.55}{\env}}}\hspace{4pt}}}}
\newcommand\cnot[1]{%
  \mathrel{\ooalign{\hfil$#1$\hfil\cr\hfil$/$\hfil\cr}}}
\newcommand\notreaches{%
  \mathrel{\ooalign{\hss$\cnot\rightsquigarrow$\hss\kern-1.45ex\raise1.0ex\hbox{{\scaleobj{0.55}{\env}}}\hspace{4pt}}}}

\newcommand\alwaysreach{%
  \mathrel{\ooalign{\hss$\square$\hss\kern-0.22ex\hbox{{$\reaches$}}}}}

\newcommand\generates{%
 \mathrel{\ooalign{\hss$\vdash$\hss\kern-0.65ex\raise0.9ex\hbox{{\scaleobj{0.55}{\env}}}}}}

\def\states{\mc{S}} % Agent state.
\def\learnf{u}

\def\apolicy{\pi} % Agent state.
\def\apolicyset{\Pi} % Agent state.

\usepackage{calligra}
\DeclareMathAlphabet{\mathcalligra}{T1}{calligra}{m}{n}
\def\valuef{v}

\def\histories{\mc{H}}
\def\rhistories{\bar{\histories}}
\def\rsuffhistories{\acute{\histories}}

\def\realizablehistabbr{\overline{\histories}}
\def\historiesastate{\histories^\circ}

\newtheorem{theorem}{Theorem}
\newtheorem*{theorem*}{Theorem}

\newtheorem{remark}[theorem]{Remark}
\newtheorem{lemma}[theorem]{Lemma}
\newtheorem{proposition}[theorem]{Proposition}
\newtheorem{definition}{Definition}
\newtheorem*{definition*}{Definition}

\numberwithin{equation}{section}
\numberwithin{theorem}{section}
\numberwithin{definition}{section}
\numberwithin{conjecture}{section}
\newmdenv[
  topline=false,
  bottomline=false,
  rightline = false,
  leftmargin=8pt,
  rightmargin=0pt,
  skipabove=16pt, % top margin
  innertopmargin=0pt,
  innerbottommargin=0pt
]{innerproof}

\newenvironment{dproof}[1][Proof]{\begin{proof}[\textbf{\textit{Proof of #1.}}] \text{\vspace{1mm}} \begin{innerproof}}{\end{innerproof}\end{proof}\vspace{1mm}}

\usepackage{etoolbox}
\makeatletter
\patchcmd{\NAT@test}{\else \NAT@nm}{\else \NAT@hyper@{\NAT@nm}}{}{}
\makeatother

\newtheorem{innercustomthm}{Theorem}
\newenvironment{customthm}[1]
  {\renewcommand\theinnercustomthm{#1}\innercustomthm}
  {\endinnercustomthm}
\newtheorem{innercustomrem}{Remark}
\newenvironment{customrem}[1]
  {\renewcommand\theinnercustomrem{#1}\innercustomrem}
  {\endinnercustomrem}
\newtheorem{innercustomprop}{Proposition}
\newenvironment{customprop}[1]
  {\renewcommand\theinnercustomprop{#1}\innercustomprop}
  {\endinnercustomprop}
\newtheorem{innercustomlem}{Lemma}
\newenvironment{customlem}[1]
  {\renewcommand\theinnercustomlem{#1}\innercustomlem}
  {\endinnercustomlem}
  
\usepackage{tikz}

\usepackage[textsize=tiny]{todonotes}

\newcommand{\creflink}[1]{\hyperref[#1]{\textcolor{blue}{\cref{#1}}}}

\usepackage{newpxmath}

\title{On the Convergence of Bounded Agents}

\author{\name David Abel \email dmabel@deepmind.com \\ \addr DeepMind
        \AND
        \name Andr{\'e} Barreto \email andrebarreto@deepmind.com \\ \addr DeepMind
        \AND
        \name Hado van Hasselt \email hado@deepmind.com \\ \addr DeepMind
        \AND
        \name Benjamin Van Roy\email benvanroy@deepmind.com \\ \addr DeepMind
        \AND
        \name Doina Precup \email doinap@deepmind.com \\ \addr DeepMind
        \AND
        \name Satinder Singh \email baveja@deepmind.com \\ \addr DeepMind}

\begin{document}

\maketitle

% --- Abstract ---
\begin{abstract}
When has an agent converged?
Standard models of the reinforcement learning problem give rise to a straightforward definition of convergence: An agent converges when its behavior or performance in each environment state stops changing.
However, as we shift the focus of our learning problem from the environment's state to the agent's state, the concept of an agent's convergence becomes significantly less clear.
%
%
% This paper.
In this paper, we propose two complementary accounts of agent convergence in a framing of the reinforcement learning problem that centers around bounded agents.
%
%
% First: performance.
The first view says that a bounded agent has converged when the minimal number of states needed to describe the agent's future behavior cannot decrease.
%
%
% Second: behavior.
The second view says that a bounded agent has converged just when the agent's performance only changes if the agent's internal state changes.
%
%
% Relationship of two proposals.
We establish basic properties of these two definitions, show that they accommodate typical views of convergence in standard settings, and prove several facts about their nature and relationship.
%
%
% Impact.
We take these perspectives, definitions, and analysis to bring clarity to a central idea of the field.
\end{abstract}

% ------------------
% -- Introduction --
% ------------------
\section{Introduction}

% AI, agents, MDPs, convergence.
%
The study of artificial intelligence (AI) is centered around agents. In reinforcement learning (RL, \citeauthor{kaelbling1996reinforcement}, \citeyear{kaelbling1996reinforcement}; \citeauthor{sutton2018reinforcement}, \citeyear{sutton2018reinforcement}), focus has traditionally concentrated on agents that interact with a Markov decision process  (MDP, \citeauthor{bellman1957markovian}, \citeyear{bellman1957markovian}; \citeauthor{puterman2014markov}, \citeyear{puterman2014markov}) or a partially observable MDP (POMDP, \citeauthor{cassandra1994acting}, \citeyear{cassandra1994acting}). Effective agents are often viewed as those that can efficiently converge to optimal behavior \citep{puterman1979convergence,kearns1998finite,szepesvari1997asymptotic}, minimize regret \citep{auer2008near,jaksch2010near,azar2017minimax}, or make a small number of mistakes before identifying a near-optimal behavior \citep{fiechter1994efficient,kearns2002near,brafman2002r,Strehl2009}. Indeed, proving that a particular learning algorithm (quickly) \textit{converges} in MDPs has long stood as a desirable property of RL agents \citep{watkins1992q,bradtke1996linear,singh2000convergence,gordon2000reinforcement}. This goal has shaped many aspects of the RL research agenda: We are often most interested in understanding whether---or how quickly---various learning algorithms converge, such as policy gradient methods \citep{sutton1999policy,singh2000convergence,fazel2018global}. In this sense, convergence plays a central role to our understanding of learning itself, and has served as a guiding principle in the design and analysis of agents.

% Convergence is straightforward in problems we care about.
%
In the settings we have tended to study it is easy to think about convergence. For example, when the problem of interest is described by an MDP or POMDP, we can think of an agent's convergence in terms of the series of functions that vary according to the environment's state, such as the agent's $Q$-function \citep{watkins1992q,majeed2018q}.

% Agent-centric RL.
%
% (i) Why agent-centric?
\paragraph{Agents and RL.} However, RL has historically been committed to a modeling imbalance: Despite the iconic two box diagram of the RL problem,\footnote{See, for example, Figure 1 of the RL survey \citep{kaelbling1996reinforcement} or Figure 3.1 of the RL textbook \citep{sutton2018reinforcement}.} our notation and modeling tools tend to focus primarily on the environment, while largely ignoring the agent. For instance, when we talk about RL in terms of an agent interacting with an MDP, we begin by fully defining the contents of the MDP. As a consequence, we orient our thinking and research pathways around the MDP, rather than the agent: We might study what happens when structure is present in the MDP's reward or transition function  \citep{bradtke1996linear,calandriello2014sparse,jin2020provably,icarte2022reward}, or in the MDP's state space \citep{kearns1999efficient,sallans2004reinforcement,diuk2008object}. However, because MDPs rely on variables associated with the environment, they do not emphasize the conceptual tools for scrutinizing agents, despite the central role of agency in AI. We suggest that, to fully understand intelligent agents, it is useful to also consider a version of RL in which attention is on the agent, rather than the environment.

% (ii) What is agent-centric?
%
To these ends, we draw inspiration from \citet{dong2022simple}, \citet{lu2021reinforcement}, \citet{konidaris2006framework}, and the work on universal AI by \citet{hutter2000theory,hutter2002self,hutter2004universal}, and study a variant of the RL problem that gives up reference to environment state and introduces agent state. This view does two things. First, it removes explicit reference to the environment state space in favour of an agent state space. After all, the agent (and often us as agent designers) do not have access to environment state. Second, it allows us to emphasize the role that boundedness plays in the nature of the agents we study, as in work by \citet{ortega2011unified}. That is, we can more sharply characterize the kinds of agents we are interested in by modelling the constraints facing real agents. This view draws from the long line of literature going back to \citeauthor{simon1955behavioral}'s bounded rationality (\citeyear{simon1955behavioral}) and its kin \citep{cherniak1990minimal,todd2012ecological,lewis2014computational,griffiths2015rational}---the relationship between \textit{intelligence} and \textit{resource constraints} is indeed fundamental \citep{russell1994provably,ortega2011unified}, and one that should feature directly in the agents we study. To this end, we build around \textit{bounded agents} (\cref{def:bounded_agent}).

% This paper.
%
\paragraph{Paper Overview.} This paper inspects the concept of agent convergence in a framing of RL focused on bounded agents. We ask: Given a bounded agent interacting with an environment, what does it mean for the agent to \textit{converge} in that environment?
%
% Main results.
We propose two notions of convergence based on an agent's behavior and performance, visualized in \cref{fig:cp_visual}.
%
% First formalism.
Our first account roughly says that an agent's behavior has converged in an environment when the minimum number of states needed to produce the agent's future behavior can no longer decrease.
%
% Second formalism.
The second account roughly says that an agent's performance has converged in an environment when the agent's performance can \textit{only} change if the agent's internal state changes.
%
% Comments on both
These definitions are based on two new quantities that capture objective properties of an agent-environment pair: (i) The limiting size %of an agent in an environment
(\cref{def:limiting_size}), and (ii) The limiting distortion %of an agent in an environment 
(\cref{def:limiting_distortion}). Moreover, we prove that \textit{both} definitions accommodate standard views of convergence in traditional problem settings. We further establish connections between the two convergence types, and discuss their potential for opening new pathways to agent design and analysis.

To summarize, this paper is about the following three points:
\begin{enumerate}
    
    % Bounded agent RL is important.
    %
    \item To understand intelligent agents, it is important to study RL centered around \textit{bounded agents}.

    % Definitions in agent-centric RL.
    %
    \item In this framing of RL, it is prudent to recover definitions of our central concepts, such as convergence.

    % Two definitions of convergence.
    %
    \item We offer two definitions of the convergence of bounded agents: in behavior (\cref{def:limiting_size}), and in performance (\cref{def:limiting_distortion}), and analyse their properties.
\end{enumerate}

% -------------------
% -- Preliminaries --
% -------------------
\section{Preliminaries: Agents and Environments}
\label{sec:preliminaries}

% Concepts overview.
%
We begin by introducing key concepts and notation of RL, including \textit{agents}, \textit{environments}, and their kin. Much of our notation and conventions draw directly from recent work by \citet{dong2022simple} and \citet{lu2021reinforcement}, as well as the vast literature on \textit{general RL} \citep{lattimore2013sample,lattimore2014theory,leike2016nonparametric,cohen2019strongly,majeed2018q,majeed2021abstractions}. We draw further inspiration from the concept of \textit{agent space} proposed by \citet{konidaris2006framework} and \citet{konidaris2006autonomous,konidaris2007building}, the disssertation by \citet{ortega2011unified} that first developed a formal view of resource-constrained agency, as well as \textit{What is an agent?} by \citet{harutyunyan2020what}.

% Notation.
%
\paragraph{Notation.} Throughout, we let capital calligraphic letters denote sets ($\mc{X}$), lower case letters denote constants and functions ($x$), italic capital letters denote random variables ($X$), and blackboard capitals indicate the natural and real numbers ($\N, \R$, $\N_0 = \N \cup \{0\}$). Additionally, we let $\Delta(\mc{X})$ denote the probability simplex over the countable set $\mc{X}$. That is, the function $\rho : \mc{X} \times \mc{Y} \ra \Delta(\mc{Z})$ expresses a probability mass function $\rho(\cdot \mid x, y)$, over $\mc{Z}$, for each $x \in \mc{X}$ and $y \in \mc{Y}$. Lastly, we use $\wedge$ as logical conjunction, $\Rightarrow$ as logical implication, as well as $\forall_{x \in \mc{X}}$ and $\exists_{x \in \mc{X}}$ to express logical quantification over a set $\mc{X}$.

We begin by defining environments and related artifacts.

% Definition: Interface.
%
\begin{definition}
An agent-environment \textbf{interface} is a pair, $(\actions, \observations)$, containing finite sets $\actions$ and $\observations$.
\end{definition}

% Elements of the interface.
%
We refer to elements of $\actions$ as \textit{actions}, denoted $a$, and elements of $\observations$ as \textit{observations}, denoted $o$.

% Definition: History.
%
\begin{definition}
For each $t \in \N_0$, a \textbf{history}, $h_t = a_0 o_1 a_1 \ldots a_{t-2} o_{t-1}  \in \histories_t$, is a sequence of alternating actions and observations, $\histories_t = \left(\actions \times \observations\right)^t$, with $\histories = \bigcup_{t=0}^\infty \histories_t$ the set of all histories.
\end{definition}

We use $h$ to refer to any history in $\histories$, $|h|$ to denote the number of $(a,o)$ pairs contained in $h$, and $hh'$ to denote the history resulting from concatenating histories $h,h' \in \histories$. Lastly, we let $\histories_{n:\infty} = \bigcup_{t=n}^\infty \histories_t$ denote all histories of length $n$ or greater, for some $n \in \mathbb{N}_0$.

% Definition: Environment.
%
\begin{definition}
\label{def:environment}
An \textbf{environment} with respect to interface $(\actions, \observations)$ is a function $\environment : \histories \times \actions \ra \Delta(\observations)$.
\end{definition}

% Generality of environments.
%
Note that this model of an environment is (i) general, in that it can express the same kinds of problems modeled by an infinite-state POMDP, and (ii) emphasizes online variations of learning. While the presence of environment states and episodes are interesting special cases, we emphasize that we do not want our results or insights to be specialized to them.

% Transition.
%
Next, we consider an abstract notion of an agent that captures the mathematical way in which experience gives rise to action selection.

% Definition: Agent.
%
\begin{definition}
\label{def:agent}
An \textbf{agent} with respect to interface $(\actions, \observations)$ is a function, $\agent : \histories \ra \Delta(\actions)$.
\end{definition}

% Note on boundedness.
%
This view of an agent ignores all of the mechanisms that might \textit{produce} behavior at each history. Indeed, it is equivalent to how others define history-based policies \citep{leike2016thompson,majeed2021abstractions}. We unpack this abstraction momentarily by limiting our focus to bounded agents (\cref{def:bounded_agent}).% that operate with finite capacity.

% Abbreviations and notation.
%
We use $\allagents$ to refer to the set of all agents, $\agents$ as any non-empty set of agents, and $\environments$ to refer to the set of all environments.

% Interaction.
%
An agent can interact with any environment that is defined with respect to the same interface $(\actions, \observations)$. The interaction takes places in discrete steps in the following way: for each $t \in \N_0$ the agent outputs $a_{t} \in \actions$ to the environment, followed by the environment passing $o_{t+1} \in \observations$ to the agent. In response to $o_{t+1}$, the agent outputs an action $a_{t+1} \in \actions$, and so on. Each agent-environment pair induces a particular set of \textit{realizable histories} as follows.

% Definition: realizable histories.
%
\begin{definition}
\label{def:realizable_histories}
The \textbf{realizable histories} of a given $(\agent,\environment)$ pair define the set of histories of any length that can occur with non-zero probability,
\begin{equation}
\histories^{\agent,\env} = \rhistories = \bigcup_{t=0}^\infty \left\{h_t \in \mc{H}_t : \prod_{k=0}^{t-1} \environment(o_k \mid h_k, a_k) \agent(a_{k} \mid h_{k}) > 0\right\}.
\end{equation}
\end{definition}

Lastly, given a realizable history $h$, we will regularly refer to the realizable history \textit{suffixes}, $h'$, which, when concatenated with $h$, produce a realizable history $hh' \in \histories^{\agent,\env}$. We define this set as follows.

% Definition: Realizable History Suffixes.
%
\begin{definition}
\label{def:realizable_history_suffixes}
The \textbf{realizable history suffixes} of a given $(\agent, \env)$ pair, relative to a history prefix $h \in \histories^{\agent, \env}$, define the set of histories that, when concatenated with prefix $h$, remain realizable,
\begin{equation}
    \histories^{\agent,\env}_h = \rsuffhistories = \{ h' \in \histories : hh' \in \histories^{\agent, \env}\}.
\end{equation}
\end{definition}

% Abbreviations.
%
When clear from context, we abbreviate $\histories^{\agent,\env}$ to $\rhistories$, and $\histories^{\agent,\env}_h$ to $\rsuffhistories$.
%
% Abbreviation combos.
%
Additionally, we occasionally combine abbreviations. For instance, recall that $\histories_{t:\infty}$ denotes all histories of length $t$ or greater, and that $\histories_h^{\agent, \env}$ denotes the realizable history suffixes following $h$. We combine these two abbreviations and let $\rsuffhistories_{t:\infty}$ refer to the realizable history suffixes of length $t$ or greater, relative to a given $h$, which is obscured for brevity.

Supported by the arguments of \citet{bowling2023settling}, we assume that the agent agents goal is captured by a received scalar signal called the \textit{reward} each time step, generated by a \textit{reward function}.

% Definition: Reward Function.
%
\begin{definition}
We call $r : \histories \ra \R$ a \textbf{reward function}.
\end{definition}

% Agnostic to reward treatment.
%
We remain agnostic to how precisely the reward function is implemented; it could be a function inside of the agent, or the reward signal could be passed as a special scalar as part of each observation. Such commitments will not have an impact on our framing. When we refer to an environment we will implicitly mean that a reward function has been chosen, too. We will commit to the view that agents are evaluated based on some function of their received future reward, defined as follows.

% Definition: Performance.
%
\begin{definition}
\label{def:performance}
The \textbf{performance}, $\valuef : \rhistories \times \allagents \times \environments \ra [\vmin, \vmax]$ is a bounded function for fixed but arbitrary constants $\vmin, \vmax \in \mathbb{R}$ where $\vmin < \vmax$. The function $\valuef$ expresses some statistic of the received future (random) rewards $R_t = r(H_t)$ produced by the future interaction between $\agent$ and $\environment$. We will use $\valuef(\agent, \environment)$ as a shorthand, and will also adopt $\valuef(\agent, \environment \mid h)$ to refer to the performance of $\agent$ on $\environment$ after any history $h \in \rhistories$.
\end{definition}

% Concrete instances of performance.
%
For concreteness, the reader may think of one choice of performance $v(\agent, \environment \mid h_t)$ as the average reward received by the agent,
\[
\liminf_{k \rightarrow \infty} \tfrac{1}{k}\mathbb{E}_{\agent,\environment}[R_{t} + \ldots + R_{t+k} \mid H_t = h_t],
\]
where $\mathbb{E}_{\agent, \environment}[\ \dots \mid H_t = h_t]$ denotes expectation over the stochastic process induced by $\agent$ and $\environment$ following $h_t$. Alternatively, we can measure performance based on the expected discounted reward received by the agent, $v(\agent, \environment) = \mathbb{E}_{\agent,\environment}[R_0 + \gamma R_1 + \ldots]$, where $\gamma \in [0,1)$ is a discount factor. The discussion that follows is agnostic to the choice of $v$.

% RL Problem.
%
From these basic components, we define a general form of the RL problem as follows.

% Definition: RL Problem.
%
\begin{definition}
    \label{def:rl_problem}
   The tuple $(\environment, \agents, \valuef)$ defines an instance of the \textbf{reinforcement-learning problem}:
    \begin{equation}
        \arg\max_{\agent \in \agents} \valuef(\agent, \environment).
    \end{equation}
\end{definition}

% Summary of RL problem.
%
The above definition is a general framing of RL without an explicit model of environment state: We are interested in designing agents $(\agent \in \agents)$ that can achieve high-performance ($\valuef$) on some chosen environment ($\environment \in \environments$).

% Central question.
%
Now, given an agent $\agent$ and environment $\environment$, our goal is to study the following question: \textit{What does it mean for $\agent$ to \textit{converge} in $\environment$?} To ground this question, we will next introduce bounded agents (\cref{def:bounded_agent}), 
a mechanistic, resource-constrained view of the agents we have so far introduced. Then, using this notion of a bounded agent, we formalise our intuitions around what it means for an agent to converge in Sections \ref{sec:convergence_in_behavior} and \ref{sec:convergence_in_performance}.

% --- Bounded Agents ---
\subsection{Bounded Agents}

% Bounded agents.
%
One advantage of the perspective on RL that emphasizes \textit{agents} is that it invites questions regarding the nature of the agents we are interested in. For example, all agents we build are ultimately \textit{bounded}: They are implemented in computers with finite capacity. Consequently, such agents cannot memorize arbitrary-length histories, or manipulate infinite precision real numbers. Thus, actual agents must do a minimum of two things: (1) maintain a finite summary of history that we call \textit{agent state}, since the agents are bounded, and (2) produce some behavior given this agent state, since they are agents. We define agent state as follows.

% Definition: state.
%
\begin{definition}
\label{def:state}
An \textbf{agent state}, $s \in \states$, denotes one possible configuration of a given agent.
\end{definition}

% Agent state.
%
Unless unclear from context, we simply refer to the agent state as \textit{state} throughout. Observe that the notion of state adopted here includes everything about the agent: Parameters, update rules, memory, weights, and so on. This is distinct from some other notions of state adopted in the literature, such as the \emph{environment state}, which can be interpreted as containing everything needed by the environment to compute the next observation, but not necessarily the information needed to compute the agent's next action. Using this concept, we model history-to-state mappings as follows.

% Definition: history-to-state function.
%
\begin{definition}
\label{def:history_to_state}
A \textbf{history-to-state} function, $\vec{\learnf} : \histories \ra \states$, is a mapping from each history to a state.
\end{definition}

% Agents can be decomposed int two.
%
Every agent $\agent$ can be decomposed into two functions: A mapping from histories to states and a mapping from states to distributions over actions. To see why, note that if we define $\vec{\learnf}$ as the identity function and let $\states = \histories$, we fully recover \cref{def:agent}.
%
% Bounded agent restriction.
However, the fact that the function $\vec{\learnf}$ must be computed by the agent restricts the class of functions that can be used, since bounded agents cannot retain histories of arbitrary sizes. We thus define bounded agents with respect to \textit{state-update functions} that can be incrementally computed based on the current state plus the most recent observation and action, as follows. %We call these functions \emph{state-update functions}:

% Definition: state-update.
%
\begin{definition}
\label{def:state_update}
A \textbf{state-update} function, $\learnf : \states \times \actions \times \observations \ra \states$, maintains state from experience and prior state.
\end{definition}

% State update.
%
The set of state-update functions is strictly smaller than the set of history-to-state functions~\citep{mccallum1996reinforcement,sutton2018reinforcement}. Despite this, we will occasionally make use of the $\vec{\learnf}$ notation for brevity. Then, given that the agent's state is updated through $\learnf$, the agent's behavior is produced by a policy as follows.

\begin{definition}
\label{def:policy}
An agent's \textbf{policy} is a function, $\apolicy : \states \ra \Delta(\actions)$, that produces behavior given the current agent state.
\end{definition}

Then, we formally introduce the notion of a \textit{bounded agent} as follows.

% Definition: Bounded Agent.
%
\begin{definition}
\label{def:bounded_agent}
An agent $\agent \in \allagents$ is said to be a \textbf{bounded agent} if there exists a tuple $(\states, s_0, \apolicy, \learnf)$, such that
\begin{equation}
    \agent(\cdot \mid h_t) = \apolicy(\cdot \mid s_t),\hspace{6mm}
    s_t = \begin{cases}
    \learnf(s_{t-1}, a_{t-1}, o_t)&t>0,\\
    s_0&t = 0,\\
    \end{cases}
\end{equation}
for all $t \in \mathbb{N}_0, h_t \in \histories$, where $\states$ is a finite agent state space, $s_0 \in \states$ is the starting agent state, $\apolicy$ is a policy, and $\learnf$ is a state-update function.
\end{definition}

% A note on size.
%
We use $|\agent|$ to refer to the size of a bounded agent's state space. For instance, a bounded agent defined over four states $\states = \{s_0, s_1, s_2, s_3\}$ has size four, and thus, $|\agent| = |\states| = 4$. Henceforth, when we refer to an agent $\agent$, we mean a bounded agent.

% Summary of fungibility.
%
This decomposition of a bounded agent is \textit{general} in the sense that it can nearly capture any agent defined over a state space of a given size. The only caveat is that we have required the state-update to be deterministic, which is in fact less expressive than a stochastic state-update. Although it may be convenient to \emph{think} of an agent as computing several functions---such as a function that updates the policy over time---all of this structure can be folded in the state-update function. This means that any bounded agent without access to random bits can be described in terms of \cref{def:bounded_agent}, including all agents that have been implemented to date such as DQN \citep{mnih2015human} and MuZero \citep{schrittwieser2020mastering}. While this may not be surprising, it does provide a simple formalism to analyze bounded agents of arbitrary complexity. %From here on, when we refer to an \textit{agent}, we mean a bounded agent.

% --------------------------------
% -- Convergence in Performance --
% --------------------------------
\subsection{The Convergence of Bounded Agents}

% % Figure: Cartoon of ct, deltat over time.
% %
\begin{figure*}[t!]
    \centering
    
    % Minimal size.
    \subfigure[Minimal Size]{\includegraphics[width=0.36\textwidth]{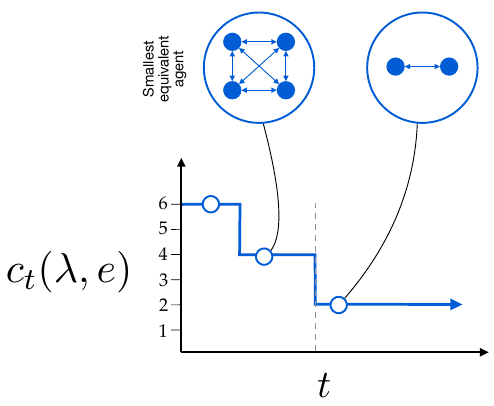}} \hspace{14mm}
    %
    % Minimal distortion.
    \subfigure[Minimal Distortion]{\includegraphics[width=0.50\textwidth]{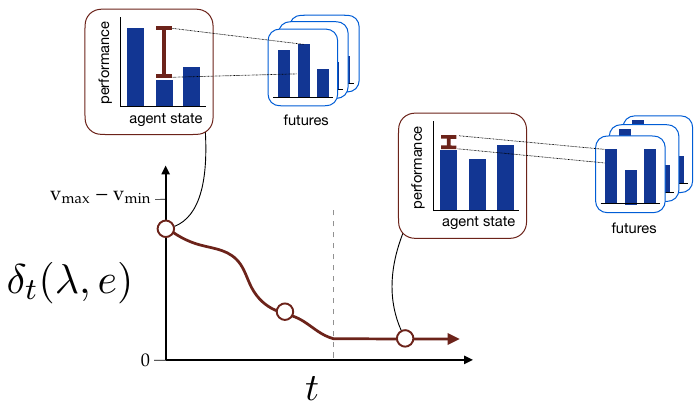}}
    
    %
    % Caption
    \caption{A cartoon visual of our two proposed definitions of convergence: (left) behavior convergence, and (right) performance convergence.  Behavior convergence tracks the sequence, $\{c_t(\agent, \environment)\}_{t=0}^\infty$, measuring the minimum number of states needed to produce the agent's behavior from the current time step across all outcomes that could occur in the environment. Performance convergence tracks the sequence, $\{\delta_t(\agent, \environment)\}_{t=0}^\infty$, measuring the maximum change in the agent's performance across return visits to its agent states. The vertical dashed line indicates the time of convergence in each case: On the left, the agent's minimal size has stopped decreasing, and on the right, the agent's distortion has stopped decreasing.}
    %
    % Label
    \label{fig:cp_visual}
\end{figure*}

% Broad overview of convergence in RL/AI.
%
The notion of convergence underlies many fundamental concepts in RL, and is part of the scientific community's parlance. But what do we \emph{mean} when we say that an agent has converged?

% Convergence more broadly.
%
When we say some object that evolves over time converges, we tend to mean that we can characterize the evolution of the object as approaching a specific point. For instance, when we speak about the convergence results of Q-learning by \citet{watkins1992q}, we mean that the limiting sequence of Q-values maintained by the algorithm arrives at a fixed point with probability one. Indeed, this is borrowing directly from the well-defined notions of the convergence of a sequence: One classical definition says roughly that a sequence of random variables, $X_1, X_2, \ldots$ converges to a number $z \in \R$ just when, for all $\epsilon \in \R_{> 0}$, with probability one there exists a $t$ such that for all $n > t$, it holds that $|z - X_{n}| < \epsilon$. Naturally, there is a plurality of such notions, such as pointwise convergence or convergence in probability. We can apply these notions to any sequence generated by the interaction of the agent with the environment. % from stochastic convergence \citep{pollard2012convergence} to analysis \citep{abbott2001understanding}

% Typical models of RL: convergence.
%
In typical models of the RL problem such as $k$-armed bandits or MDPs, the choice of the problem formulation immediately restricts our focus to sequences that are suggested by the presence of environment state: For instance, we might consider the agent's choice of an action distribution over time (in the case of bandits, when there is only one environment state) or the agent's choice of a policy over time (in the case of an MDP). We can imagine generalizing to POMDPs by simply asking whether the agent's behavior or performance has converged relative to the POMDP's hidden state. However, from the view focused on agents, it is less clear which sequence we might be interested in.

Ultimately, we will introduce two new fundamental definitions of an agent-environment pair that induce sequences (pictured in \cref{fig:cp_visual}), whose limits reflect the convergence of a bounded agent from two perspectives:
\begin{enumerate}
    \item \textit{(Behavior)} Minimal Size, $c_t(\agent,\environment)$, given by \cref{def:size_time_t}: A measurement of the smallest state space needed to produce the agent's behavior across all realizable futures in the environment.
    
     \item \textit{(Performance)} Distortion, $\delta_t(\agent,\environment)$, given by \cref{def:distortion_time_t}: A measurement of the gap in the agent's performance across future visits to the same agent state in the environment.
\end{enumerate}

We next motivate and define each of these quantities in more detail. Our analysis reveals that the limit of each sequence exists for every $(\agent, \environment)$ pair (\cref{thm:ct_properties}, \cref{thm:delta_t_properties}). In MDPs, we show these two measures are connected (\cref{prop:bandmdp_conv_implies_bconv}, \cref{prop:bandmdp_conv_implies_pconv}), which may explain why, historically, there was less pressure to draw the distinction between convergence in performance and behavior. However, in the most general settings, we show they are in fact distinct (\cref{prop:conv_time_diff}).

% -----------------------------
% -- Convergence in Behavior --
% -----------------------------
\section{Convergence in Behavior}
\label{sec:convergence_in_behavior}

We first explore convergence from the viewpoint of \textit{behavior}. Intuitively, we might think an agent has converged in a bandit problem when the agent only pulls a single arm forever after. How might we generalize this intuition?

% --- In General Environments ---
\subsection{Limiting Size: $c_\infty(\agent, \environment)$}

% Single sentence summary.
%
Our first definition is built around the following perspective: An agent has converged in behavior in an environment when the minimal number of states needed to describe the agent's future behaviors can no longer decrease. We focus around \textit{uniform} convergence, in which agents converge in an environment uniformly across all realizable histories $\rhistories$, but note that alternative formulations are possible (and can be easily derived from ours).

% Intuition.
%
To define this idea carefully, we introduce one new fundamental concept of an agent-environment pair: The number of states needed, in the limit, to reproduce the agent's future behavior in the environment. First, let
\begin{equation}
    \agents_n = \{\agent \in \allagents : |\agent| = n\},
\end{equation}
denote the set of agents operating over a state space of size $n \in \N$. We define an agent's \textit{minimal size} at time $t\in \N_0$ in terms of the size of the smallest agent that produces identical future behavior to $\agent$ in $\environment$ as follows.

% Definition: Minimal size at time t.
%
\begin{definition}
\label{def:size_time_t}
The \textbf{minimal size from time $\bm{t}$} of agent $\agent$ in environment $\environment$ is denoted
\begin{equation*}
    c_t(\agent, \environment) =  \min \{n \in \mathbb{N} : \forall_{h \in \rhistories_{t:\infty}} \exists_{\agent_n \in \agents_n} \forall_{h' \in \rsuffhistories}\ \agent(hh') = \agent_n(hh')\}.
\end{equation*}
\end{definition}

% Intuitive description of the above definition.
%
So, given any time $t \in \N_0$, we measure the minimal number of states needed to reproduce the agent's behavior \textit{forever after} in the environment. Intuitively, this number describes how compressed the agent \textit{could be}, while still producing the same behavior in the current environment from the current time on. Notice that $c_t(\agent, \environment) \leq |\agent|$, as the agent cannot be both (i) minimal and, (ii) larger than its current size. Further observe that $c_t(\agent, \environment) \geq 1$ by definition---we suppose the smallest agent has a single state by convention.

The minimal size from time $t$ produces a sequence, $\{c_t(\agent, \environment)\}_{t=0}^\infty$, with which we can sensibly discuss the convergence of an agent. In particular, the limit of this sequence captures the minimal number of states needed to describe the agent's behavior in the limit of interaction with $\environment$.%Formally, this is defined as follows.

% Definition: limiting size.
%
\begin{definition}
\label{def:limiting_size}
The \textbf{limiting size} of agent $\agent$ interacting with environment $\environment$ is defined as
\begin{equation}
    c_{\infty}(\agent, \environment) = \lim_{t \ra \infty} c_t(\agent, \environment).
\end{equation}
\end{definition}

% Brief description.
%
We take the value of this limit to be capture to capture what it means for an agent's behavior to converge. To see why, we next turn to the analysis of the limiting size.

% --- Analysis: Behavior Convergence ---
\subsection{Analysis: Behavior Convergence}

The limiting size comes with several useful properties that strengthen its case as a convergence definition. All proofs are presented in the Appendix.% mirroring those of \cref{thm:delta_t_properties}.

% Theorem: this sequence is monotonic.
%
\begin{theorem}
\label{thm:ct_properties}
For every $(\agent, \environment)$ pair:
\begin{enumerate}[(a)]
    %
    % Non-increase.
    \item The sequence $\{c_t(\agent, \environment)\}_{t=0}^\infty$ is non-increasing.
    \item $1 \leq c_t(\agent, \environment) \leq |\agent|, \forall_{t \geq 0}$.
    %
    %
    % Limit exists.
    \item $c_{\infty}(\agent, \environment)$ exists.
\end{enumerate}
\end{theorem}

% Summary of the measure.
%
Hence, by point (c), we can sensibly talk about any agent's limiting size in any given environment.

We further point out several special cases---when an agent is equivalent to a memoryless policy or a fixed $k$-th order policy, its limiting size must obey certain bounds. To be precise, we let,
\begin{equation}
    \Pi_{\mc{O}} = \{ \pi : \observations \ra \Delta(\actions)\},
\end{equation}
denote the set of all \textit{memoryless} policies.

% Corollary
\begin{remark}
\label{rem:ct_policies}
For every $(\agent, \environment)$ pair:
\begin{enumerate}[(a)]
    %
    % Memoryless policy.
    \item If $\agent$ is equivalent to a memoryless policy, $\apolicy_\mc{O} \in \apolicyset_{\mc{O}}$, then $c_\infty(\agent, \environment) \leq |\observations|$, in every environment. 
    %
    % k-th order policy?
    %
    \item If $\agent$ is equivalent to a $k$-th order policy, $\apolicy : \observations^k \ra \Delta(\actions)$, then $c_\infty(\agent, \environment) \leq |\observations|^{k}$, in every environment. 
\end{enumerate}
\end{remark}

This remark notes that, when an agent is equivalent to a memoryless policy, its limiting size will be upper bounded by the size of the observation space. More specifically, we next establish that the limiting size accommodates at least one simple view of convergence in bandits and MDPs in which an agent's behavior eventually becomes uniformly equivalent to a distribution over actions (in bandits) or a memoryless policy (in MDPs).

% Theorem: beta-convergence and bandits/MDPs.
%
\begin{proposition}
\label{prop:bandmdp_conv_implies_bconv}
The following two statements hold:
\begin{enumerate}[(a)]

    % Bandits.
    %
    \item (Bandits) If an agent $\agent$ uniformly converges in a $k$-armed bandit $\environment$ in the sense that,
     \begin{equation}
        \exists_{t \in \mathbb{N}_0} \exists_{\rho \in \Delta(\actions)} \forall_{h \in \rhistories_{t:\infty}}\ \rho = \agent(h),
    \end{equation}
    then $c_\infty(\agent, \environment) = 1$.
    
    % MDPs.
    %
    \item (MDPs) If an agent $\agent$ uniformly converges in an MDP $\environment$ in the sense that,
    \begin{equation}
        \exists_{t \in \mathbb{N}_0} \exists_{\pi_\observations \in \Pi_\observations} \forall_{hao \in \rhistories_{t:\infty}}\ \pi_\observations(o) = \agent(hao),
        \label{eq:mdp_converge_policy}
    \end{equation}
    then $c_\infty(\agent, \environment) \leq |\observations|$.
\end{enumerate}
\end{proposition}

% Commentary.
%
Thus, an agent's minimal limiting size has an intuitive value in well-studied settings: In a bandit, a convergent agent eventually becomes a one-state agent; in an MDP, a convergent agent eventually becomes (at most) a $|\mc{O}|$-state agent. It is an open (and perhaps unanswerable) question as to what canonical agents converge to in more general environments, but the measure $c_\infty(\agent, \environment)$ can give us some insight.

% Section summary.
%
To summarize, we have introduced an agent's \textit{limiting size} as a quantity that reflects the convergence of an agent's behavior in a given environment. In bandits and MDPs, the measure indicates the agent converges to a particular choice of distribution over actions, and a mapping from each environment state to a distribution over actions, respectively. In general, an agent's behavior converges when its minimal size stops decreasing.

% --- Convergence in Performance ---
\section{Convergence in Performance}
\label{sec:convergence_in_performance}

% Overview of performance view.
%
We next explore the convergence of an agent's performance, as studied in prior work in general RL \citep{lattimore2011asymptotically,lattimore2014theory,leike2016thompson}. Indeed, we are often interested in an agent's asymptotic performance, or its rate of convergence to the asymptote \citep{fiechter1994efficient,szepesvari1997asymptotic,kearns1998finite}. As we inspect learning curves in practice, it is common to say an agent has converged when the learning curve levels off. Moreover, reward or performance sequences induce a series of numbers over time, and thus can be readily mapped to classical accounts of the convergence of sequences.

% --- In General Environments ---
\subsection{Limiting Distortion: $\delta_\infty(\agent, \environment)$}

% Beyond MDPs: without env state, convergence is unclear.
%
Recall that in an environment $\environment$ there is no explicit reference to state---only histories. How might we think of an agent's performance converging? We could consider a strict approach that says an agent's performance converges when it is constant across \textit{all} future histories. This is much too strong, however, as it fails to accommodate relevant changes in performance that occur solely due to the change in history. For example, in the case that $\environment$ can be captured by an MDP (and thus, the observation is a sufficient statistic of reward, transitions, and performance), then surely we want to allow changes in performance as the agent changes between environment state. What happens when we no longer have an explicit reference to environment state?

% Structure from bounded agents.
%
As with the behavioral view, we find that bounded agents provide the needed structure---each bounded agent is comprised of a finite state space, $\states$. We can then define performance convergence relative to agent state by inspecting whether the agent's performance changes across return visits to each of its own agent states. Intuitively, if the agent's performance is the same every time it returns to the same agent state, then we say the agent has converged. We formally capture this notion through the \textit{limiting distortion} of an agent-environment pair, which serves as the basis for our second definition of agent convergence.

% Notation: history pairs where agent revisits the same agent state.
%
To introduce this concept, we require extra notation to capture return visits to the same agent state. We are interested in realizable histories that can be subdivided, $hh'$, such that the agent maps $h$ and $hh'$ to the same internal agent state. We denote this set
\begin{align}
    \label{eq:hcirc}
    \historiesastate_t =\ \{(h, h') \in \histories :\ hh' \in \rhistories_{t:\infty}\ &\wedge\ |h| \geq t,\\
    &\wedge\ |h'| > 0, \nonumber \\
    &\wedge\ \vec{\learnf}(h) = \vec{\learnf}(hh')\}. \nonumber
\end{align}

% Intuition.
%
Note that for \textit{all} bounded agents, this set is non-empty for all environments $\environment \in \environments$ and all times $t \in \mathbb{N}_0$.

% Lemma.
%
\begin{lemma}
\label{lem:hcirc_non_empty}
$|\mc{H}_t^\circ| > 0$ for any $(\agent, \environment)$ pair and time $t \in \mathbb{N}_0$.
\end{lemma}

% Distortion overview.
%
We can then consider the gap in the agent's performance conditioned on the fact that the agent occupies the same agent state at two different times. This quantity, called \textit{distortion}, analysed recently by \citeauthor{dong2022simple} under a more detailed decomposition of agent state (\citeyear{dong2022simple}, Equations 5 \& 6), is closely related to the degree of value error in aggregating states \citep{van2006performance,li2006towards} or histories \citep{majeed2018q,majeed2019performance,majeed2021abstractions}. We adapt these quantities in our notation as follows.

% Definition: distortion.
%
\begin{definition}
\label{def:distortion_time_t}
The \textbf{distortion from time $\bm{t}$} of $\agent$ in $\environment$ is
\begin{equation}
\label{eq:distortion_time_t}
    \delta_t(\agent,\environment) = \\
    \sup_{(h,h') \in \historiesastate_t}
    |\valuef(\agent, \environment \mid h) - \valuef(\agent, \environment \mid hh')|.% \leq d.
\end{equation}
\end{definition}

% Example.
%
Intuitively, the distortion of an agent expresses the largest gap in performance across situations in which the agent is at the same agent state. An agent with high distortion is one in which the agent will produce very different performance as it returns to the same agent state. On the other extreme, an agent with zero distortion is one in which the performance is constant every time the agent is at the same agent state. In this way, the distortion captures whether the agent-environment interaction is non-stationary in a way that is relevant to performance.

Using this measure, we introduce the limiting distortion, the central definition of this section.

% Definition: Limiting distortion.
%
\begin{definition}
\label{def:limiting_distortion}
The \textbf{limiting distortion} of $\agent$ in $\environment$ is
\begin{equation}
    \delta_\infty(\agent, \environment) = \lim_{t \ra \infty} \delta_t(\agent, \environment).
\end{equation}
\end{definition}

% Intuition.
%
Roughly, $\delta_\infty$ measures how well an agent's state can predict the agent's own performance across all subsequent histories. This quantity will be zero just when, every time the agent visits one of its states, the factors of the environment that determine the agent's performance are the same. Thus, the limiting distortion measures the degree of performance-relevant non-stationarity present in the indefinite interaction with the environment.

% --- Performance Convergence ---
\subsection{Analysis: Performance Convergence}

To motivate this quantity as a meaningful reflection of an agent's performance convergence, we show that the limiting distortion has the following properties.

% Theorem: Properties of delta_t
%
\begin{theorem}
\label{thm:delta_t_properties}
For every $(\agent, \environment)$ pair:
\begin{enumerate}[(a)]
    \item The sequence $\{\delta_t(\agent, \environment)\}_{t=0}^\infty$ is non-increasing.
    
    \item $0 \leq \delta_t(\agent, \environment) \leq (\vmax - \vmin), \forall_{t \geq 0}$.
    
    \item $\delta_\infty(\agent, \environment)$ exists.
\end{enumerate}
\end{theorem}

% Commentary.
%
Crucially, property (c) of \cref{thm:delta_t_properties} shows that the limiting distortion exists for every $(\agent, \environment)$ pair. The relevant questions, then, are (i) the rate at which the agent reaches $\delta_\infty(\agent,\environment)$ , and (ii) what the value $\delta_\infty(\agent, \environment)$ actually is. We suggest that differing values of $\delta_\infty(\agent, \environment)$ indicate a meaningful difference in the interaction stream produced by $\agent$ and $\environment$---and thus, reflect the convergence of $\agent$ in $\environment$. We motivate this intuition with the following results.

In standard settings, behavior and performance convergence \textit{tend} to co-occur. That is, in an MDP, given an agent whose behavior converges in the sense of \cref{prop:bandmdp_conv_implies_bconv}, that same agent's performance also converges, as follows. %We make this concrete in the following result.

% Proposition: performance convergence and bandits/MDPs.
%
\begin{proposition}
\label{prop:bandmdp_conv_implies_pconv}
The following two statements hold:
\begin{enumerate}[(a)]

    % Bandits.
    %
    \item (Bandits) If an agent $\agent$ uniformly converges in a $k$-armed bandit $\environment$ in the sense that,
     \begin{equation}
        \exists_{t \in \mathbb{N}_0} \exists_{\rho \in \Delta(\actions)} \forall_{h \in \rhistories_{t:\infty}}\ \rho = \agent(h),
    \end{equation}
    then $\delta_\infty(\agent, \environment) = 0$.
    
    % MDPs.
    %
    \item (MDPs) If an agent $\agent$ uniformly converges in an MDP $\environment$ in the sense that,
    \begin{equation}
        \exists_{t \in \mathbb{N}_0} \exists_{\pi_\observations \in \Pi_\observations} \forall_{hao \in \rhistories_{t:\infty}}\ \pi_\observations(o) = \agent(hao),
        \label{eq:mdp_converge_policy2}
    \end{equation}
    then $\delta_\infty(\agent, \environment) = 0$.
\end{enumerate}
\end{proposition}

% Performance convergence in MDPs.
%
This result mirrors that of \cref{prop:bandmdp_conv_implies_bconv} on the behavioral side. Recall, however, that the style of convergence characterized by \cref{eq:mdp_converge_policy} is restrictive, and will not necessarily accommodate the style of convergence of, say, tabular Q-learning; for many stochastic environments it is unlikely that there is a finite $t$ at which Q-learning is equivalent to a memoryless policy uniformly over all histories. To this end, we next provide a general result about the performance convergence of agents with a particular form.

% Proposition: Agent state memorization is sufficient.
\begin{proposition}
\label{prop:delta_is_0_mdp_astate}
Any $\agent$ and MDP $e$ that satisfy
\begin{equation}
    \label{eq:mdp_astate_delta_condition}
    \learnf(s,a,o) = \learnf(s',a',o')\ \Rightarrow\ o = o',\hspace{8mm} \forall_{s,s'\in\states}, \forall_{a,a'\in \actions}, \forall_{o,o'\in\observations},
\end{equation}
ensure $\delta_t(\agent, \environment) = 0$ for all $t \in \mathbb{N}_0$, and thus, $\delta_\infty(\agent, \environment) = 0$.
\end{proposition}

Note that the condition of \cref{eq:mdp_astate_delta_condition} is satisfied for all bounded agents that always store the most recent observation in memory, as is typical of many agents we implement. In this sense, bounded implementations of Q-learning may ensure $\delta_\infty(\agent, \environment) = 0$. It is an open question as to whether we can design a version of bounded Q-learning that ensures convergence in MDPs in the sense of the main result of \citet{watkins1992q}. For instance, it is unclear how to design an appropriate step-size annealing schedule with only finitely many agent states.

Conversely, when we consider the same exact kind of environment but impose further constraints on the agent, we find that the limiting distortion can be greater than zero.

% Proposition: Constant step-size Q-learning.
%
\begin{proposition}
\label{prop:pconv_nonstat_env}
There is a choice of MDP $\environment$ and bounded Q-learning $\agent$ that yields $\delta_\infty(\agent, \environment) > 0$.
\end{proposition}

The contrast of \cref{prop:delta_is_0_mdp_astate} with \cref{prop:pconv_nonstat_env} draws a distinction between the convergence of different agent-environment pairs. In particular, when the environment models an MDP, we saw that bounded agents that memorize the last observation satisfy $\delta_\infty(\agent, \environment) = 0$. However, when we consider a Q-learning agent with limitations to its representational capacity, we find it may be the case that $\delta_\infty(\agent, \environment) > 0$.

Lastly, we recall that by \cref{prop:bandmdp_conv_implies_bconv} and \cref{prop:bandmdp_conv_implies_pconv}, behavioral and performance convergence tend to co-occur in traditional problems like MDPs.  As a final point, we show that, in general environments, there are $(\agent, \environment)$ pairs for which the time of behavior convergence is different from the time of performance convergence, thereby providing some formal support for treating these two views as distinct.

% Prop: Convergence time differs.
%
\begin{proposition}
\label{prop:conv_time_diff}
For any $(\agent, \environment)$ pair and any choice of $\beta, \epsilon \in \mathbb{R}_{\geq 0}$ let
\begin{align*}
    % Behavior
    %
    t_\beta(\agent, \environment) &= \min \{t \in \mathbb{N}_0 : |c_t(\agent, \environment) - c_\infty(\agent, \environment)| \leq \beta\}, \\
    %
    % Performance
    t_\epsilon(\agent,\environment) &= \min \{t \in \mathbb{N}_0 : |\delta_t(\agent, \environment) - \delta_\infty(\agent, \environment)| \leq \epsilon\}.
\end{align*}
Then, for $\beta=0$, $\epsilon=0$, there exists pairs $(\agent, \environment)$ and $(\agent', \environment')$ such that
    \begin{equation}
        t_\epsilon(\agent,\environment) < t_\beta(\agent, \environment),
        \hspace{4mm}
        t_\beta(\agent',\environment') < t_\epsilon(\agent', \environment').
    \end{equation}
\end{proposition}

This result indicates that, in general, even when an agent's limiting distortion has been reached, it does not mean that its limiting size has been reached, and vice versa. This tells a partial story as to why it is prudent to draw the concept of convergence in behavior and performance apart in more general environments. It is further worth noting that, in some environments, we suspect it is possible that $t_\epsilon$ may not exist, though $t_\beta$ is guaranteed to exist by definition of $c_\infty$.

% ----------------
% -- Conclusion --
% ----------------
\section{Conclusion}
\label{sec:conclusion}

% Overview.
%
The notion of \textit{convergence} is central to many aspects of agency---for example, we are often interested in designing agents that converge to some fixed high-quality behavior. Similarly, the notion of \textit{learning} is intimately connected to how we think about convergence: An agent that eventually converges is, in many cases, one that stops learning.
%
% Detailed paper summary.
%
In this paper we have presented a careful examination of the concept of convergence for bounded agents in an of RL that emphasizes \textit{agents} rather than environments. We explored two perspectives on how to think about the convergence of a bounded agent operating in a general environment: from behavior  (\cref{def:limiting_size}), and from performance (\cref{def:limiting_distortion}). We established simple properties of both formalisms, proved that they reflect some standard notions of convergence, and bear interesting relation to one another. We take these formalisms, definitions, and results to bring new clarity to a central concept of RL, and hope that this work can help us to better design, analyse, and understand learning agents.

% What might this do for us?
%
The perspectives and definitions here introduced suggest a number of new pathways for the analysis and design of agents. For instance, both the limiting size and distortion of an agent can be useful guides for developing or evaluating the learning mechanisms that drive our agents; in some domains, we may explicitly \textit{want} to build agents that converge in either sense, or to evaluate agents in environments that \textit{require} a larger limiting size or distortion from its agents. Similarly, these quantities may suggest new ways to sharpen the problem we are ultimately interested in by focusing only on agents that have a minimal limiting size or distortion---as is likely the case of all agents of interest. Additionally, in future work, it would be useful to develop efficient algorithms that can \textit{estimate} the limiting size or distortion of an actual agent-environment pair. Lastly, we emphasize that while behavior and performance are natural choices for a conception of convergence, we do not here discuss notions of convergence based around epistemic uncertainty \citep{lu2021reinforcement}, but acknowledge its potential significance for future work.

% --- Acknowledgements ---
\subsubsection*{Acknowledgments}
The authors are grateful to Mark Rowland for comments on a draft of the paper. We would also like to thank the attendees of the 2023 Barbados RL Workshop, as well as Elliot Catt, Will Dabney, Steven Hansen, Anna Harutyunyan, Joe Marino, Joseph Modayil, Remi Munos, Brendan O'Donoghue, Matt Overlan, Tom Schaul, Yunhao Tang, Shantanu Thakoor, and Zheng Wen for inspirational conversations.

% ------------------
% -- Bibliography --
% ------------------
\bibliographystyle{tmlr}
\bibliography{main}

% --- Appendix ---
\appendix
% --- Appendix ---
\section{Appendix}

We present each result and its associated proof in full detail, beginning with proofs of results presented in \cref{sec:convergence_in_behavior}.

% Theorem: this sequence is monotonic.
%
\begin{customthm}{3.1}
For every $(\agent, \environment)$ pair, the following properties hold:
\begin{enumerate}[(a)]
    %
    % Non-increase.
    \item The sequence $\{c_t(\agent, \environment)\}_{t=0}^\infty$ is non-increasing.
    \item $1 \leq c_t(\agent, \environment) \leq |\agent|, \forall_{t \geq 0}$.
    %
    %
    % Limit exists.
    \item $c_{\infty}(\agent, \environment)$ exists.
\end{enumerate}
\end{customthm}

% Proof.
\begin{dproof}[\cref{thm:ct_properties}]
We prove each property separately.

% (a) Sequence is non-increasing.
%
\textit{(a) The sequence $\{c_t(\agent, \environment)\}_{t=0}^\infty$ is non-increasing.}

Note that the set $\realizablehistabbr_{t:\infty}$ induces a sequence of sets $\realizablehistabbr_{0:\infty}, \realizablehistabbr_{1:\infty}, \ldots$ such that
\begin{equation}
\label{proof_eq:realiz_h_subset}
\realizablehistabbr_{t+1:\infty} \subseteq \realizablehistabbr_{t:\infty},
\end{equation}
for all $t \geq 0$. The non-increasing nature of the sequence $\{c_t(\agent, \environment)\}_{t=0}^\infty$ holds by immediate consequence of this subset relation, as the $\sup$ ensures that
\[
\sup \mc{X}_{t+1} \leq \sup \mc{X}_{t} ,
\]
for countable sets $\mc{X}_{t+1}$ and $\mc{X}_t$ where $\mc{X}_{t+1} \subseteq \mc{X}_t$. Therefore, $c_{t+1}(\agent, \environment) \leq c_t(\agent,\environment)$ for all $t \geq 0$. \checkmark.

% First, notice that at each time $t$, the quantity $c_t(\agent, \environment)$ measures the number of states needed to mimic the behavior of $\agent$ over \textit{all possible futures} that could occur in $\environment$. Hence, we are taking the min over every future history that could occur with non-zero probability, and so there cannot exist a future time $t+n$, for any $n \in \N_0$, at which the agent needs \textit{more} states than $c_t(\agent, \environment)$. \checkmark

\vspace{5mm}

% 2. 1 \leq c_t \leq |agent|.
%
\textit{(b) $1 \leq c_t(\agent, \environment) \leq |\agent|,\ \forall_{t \geq 0}$.}

Second, note that the lower bound trivially holds since the smallest an agent can be is a single state (simply by convention---we could opt for the empty state space as the smallest agent, but this choice is arbitrary). Next, we note that the upper bound also holds in a straightforward way: The minimal state space size of the agent can't be larger than the agent's \textit{actual} state space. \checkmark

\vspace{5mm}
% (c) Lim always exists.
%
\textit{(c) $c_\infty(\agent, \environment)$ exists.}

% Proof.
The property follows directly by consequence of the agent's boundedness: The actual agent $\agent$ uses a state space of some finite size, $b$. Thus, for every time $t$, in every environment, there is at least one $b \in \N$, and one bounded agent that produces the same behavior as the agent: The agent itself. Therefore, by the non-increasing nature of the sequence, there must exist a point at which $c_\infty(\agent, \environment)$ stops decreasing, due to the floor at the minimum possible state-space size of one. \checkmark

\vspace{5mm}

This completes the argument for all three statements, and thus completes the proof. \qedhere
\end{dproof}

% Remark
\begin{customrem}{3.2}
For every $(\agent, \environment)$ pair:
\begin{enumerate}[(a)]
    %
    % Memoryless policy.
    \item If $\agent$ is equivalent to a memoryless policy, $\apolicy_\mc{O} \in \apolicyset_{\mc{O}}$, then $c_\infty(\agent, \environment) \leq |\observations|$, in every environment. 
    %
    % k-th order policy?
    %
    \item If $\agent$ is equivalent to a $k$-th order policy, $\apolicy : \observations^k \ra \Delta(\actions)$, then $c_\infty(\agent, \environment) \leq |\observations|^{k}$, in every environment. 
\end{enumerate}
\end{customrem}

% Proof.
\begin{dproof}[\cref{rem:ct_policies}]

While (a) is clearly a special case of (b), we provide a proof of each result for the sake of clarity.

% 2. Memoryless has b=1.
%
\textit{(a) If $\agent$ is equivalent to a memoryless policy, then $c_\infty(\agent, \environment) \leq |\observations|$, in every environment.}

To be precise, when we say an $\agent$ is equivalent to a memoryless policy, we mean there is a choice of $\pi_\observations \in \Pi_\observations$ such that,
\begin{equation}
\label{proof_eq:agent_memoryless_policy}
\agent(h) = \pi_\observations(h),\ \forall_{h \in \realizablehistabbr}.
\end{equation}

Now that this term is clear, we see why the fact holds. The agent's behavior can always be described by a process that sets its agent state to be the last observation, then acts according to the corresponding $\pi_\observations$ from \cref{proof_eq:agent_memoryless_policy}. That is, consider $\agent^\diamond$ that (1) sets $\states = \observations$, (2) sets $s$ = $o$, and (3) acts according to $\pi_\observations^\diamond$,
\[
\pi_{\observations}^\diamond(s) = \pi_{\observations}(o).
\]
Clearly this agent only requires at most $|\observations|$ states, and thus, $c_\infty(\agent,\environment) \leq |\observations|$. \checkmark

% state agent, and thus, by definition of $c_\infty(\agent, \environment)$, we find $c_0(\agent, \environment) = c_t(\agent, \environment) = c_\infty(\agent, \environment) = 1$, for all $t \in \mb{N}_0$. Note that this does not depend on the environment. \checkmark

\vspace{5mm}

% 3. k-th order has b \leq ...
%
\textit{(b) If $\agentf$ is equivalent to a $k$-th order policy, $\apolicy : \observations^k \ra \Delta(\actions)$, then $c_\infty(\agent, \environment) \leq |\observations|^k$, in every environment. }

% Proof.
The reasoning is similar to point (3.). Let us first be precise: When we say the agent $\agent$ \text{is} a $k$-th order policy, $\apolicy : \observations^k \ra \Delta(\actions)$, we mean the agent's behavior can be described by some fixed function $\apolicy^{(k)} : \observations^k \ra \Delta(\actions)$ in every environment. That is, $\agent = (\states, s_0, \apolicy, \learnf)$ operates over some finite state space, and uses a policy $\apolicy : \states \ra \Delta(\actions)$. But, this agent ensures there exists a $k$-th order policy $\apolicy^{(k)}$ such that, for any $t$, for every $h_t = a_0 o_1, \ldots, a_{t-2} o_{t-1}$,
\[
    \agent(\cdot \mid h_t) = \apolicy^{(k)}(\cdot \mid o_{t-k-1}, \ldots, o_{t-2}, o_{t-1}).
\]

We note that handling the indices for the first observations where $t < k$ must be done carefully, but the reasoning remains the same.

Hence, we can define a new agent, $\agent^\diamond = (\states^\diamond, s_0^\diamond, \apolicy^\diamond, \learnf^\diamond)$, with $\states^\diamond = \observations^{k}$, yielding:
\[
    \agent(\cdot \mid h_t) = \agent^\diamond(\cdot \mid h_t) = \apolicy^{\diamond}(\cdot \mid \underbrace{s_{t-1}^\diamond}_{o_{t-k-1}, \ldots, o_{t-1}}).
\]
By construction of $\agent^\diamond$, the most states ever needed to mimic the behavior of $\agent$ is $|\observations|^{k}$. 

We note that the relation is an \textit{inequality}, $c_\infty(\agent, \environment) \leq |\observations|^{k}$, rather than a strict equality, as in some environments not every length $k$ sequence of observations may be required to produce the original agent's behavior. \checkmark

This completes the argument for both statements. \qedhere

\end{dproof}

% Proposition.
%
\begin{customprop}{3.3}
%Convergence in MDPs and bandits implies $\beta$-convergence.
The following two statements hold:
\begin{enumerate}[(a)]
    \item (Bandits) If an agent $\agent$ converges in a $k$-armed bandit $\environment$ in the sense that,
     \begin{equation}
        \exists_{t \in \mathbb{N}_0} \exists_{\rho \in \Delta(\actions)} \forall_{h \in \realizablehistabbr_{t:\infty}}\ \rho = \agent(h),
    \end{equation}
    then $c_\infty(\agent, \environment) = 1$.%the agent $\beta$-converges in $\environment$, for $\beta = 1$.
    
    \item (MDPs) If an agent $\agent$ converges in an MDP $\environment$ in the sense that,
    \begin{equation}
        \exists_{t \in \mathbb{N}_0} \exists_{\pi_\observations \in \Pi_\observations} \forall_{hao \in \realizablehistabbr_{t:\infty}}\ \pi_\observations(o) = \agent(hao),
        %\label{eq:mdp_converge_policy}
    \end{equation}
    then $c_\infty(\agent, \environment) \leq |\observations|$.%the agent $\beta$-converges in $\environment$, for some $\beta\leq |\observations|$.
\end{enumerate}
\end{customprop}

% Proof.
%
\begin{dproof}[\cref{prop:bandmdp_conv_implies_bconv}]
We prove the two properties separately.

\textit{(a) (Bandits) If an agent $\agent$ converges in a $k$-armed bandit $\environment$ in the sense that,
    \[
        \exists_{t \in \mathbb{N}_0} \exists_{\rho \in \Delta(\actions)} \forall_{h \in \realizablehistabbr_{t:\infty}}\ \rho = \agent(h),
    \]
    then $c_\infty(\agent, \environment) = 1$.}
    
We know by assumption that there is a time $t$ such that the agent's chosen action distribution for all realizable histories is identical. Note that a fixed choice of action distribution can be captured by a memoryless policy $\pi_\observations \in \Pi_\observations$. Since $\environment$ is a $k$-armed bandit, we know there is only one observation, $|\observations|=1$.  Thus, by point (a) of \cref{rem:ct_policies}, we conclude $c_\infty(\agent, \environment) \leq |\observations|$, where $|\observations| = 1$. \checkmark

\textit{(b) (MDPs) If an agent $\agent$ converges in an MDP $\environment$ in the sense that,
    \[
        \exists_{t \in \mathbb{N}_0} \exists_{\pi_\observations \in \Pi_\observations} \forall_{hao \in \realizablehistabbr_{t:\infty}}\ \pi_\observations(o) = \agent(hao),
    \]
    then $c_\infty(\agent, \environment) \leq |\observations|$.}
    
The argument is similar to the previous point: We know by assumption that there is a time $t$ such that the agent's behavior in all realizable histories is equivalent to some memoryless policy, $\pi_\mc{O} \in \Pi_\mc{O}$. By point (a) of \cref{rem:ct_policies}, we know that any agent that is equivalent to such a policy ensures $c_\infty(\agent, \environment) \leq |\observations|$. \checkmark

This completes the proof of both statements, and thus concludes the argument. \qedhere
\end{dproof}

% \spacerule

%%%%%%%%%%%%%%%%%%%%%
%%%%% Section 4 %%%%%
%%%%%%%%%%%%%%%%%%%%%
\section{\cref{sec:convergence_in_performance} Proofs: Convergence in Performance}

We now present proofs of results from \cref{sec:convergence_in_performance}.

% Lemma.
%
\begin{customlem}{4.1}
$|\mc{H}_t^\circ| > 0$ for any $(\agent, \environment)$ pair and time $t \in \mathbb{N}_0$.
\end{customlem}

% Proof.
\begin{dproof}[\cref{lem:hcirc_non_empty}]
The intuition is simple to state: All bounded agents must \textit{eventually} return to at least one of their agent states.

In more detail, recall that in \cref{eq:hcirc} we define $\historiesastate_t$ to contain all history pairs $(h,h')$ such that the following four conditions hold:
\begin{enumerate}[(i)]
    \item $hh' \in \realizablehistabbr_{t:\infty}$,
    
    \item $|h| \geq t$,
    
    \item $|h| > 0$.
        
    \item $\vec{\learnf}(h) = \vec{\learnf}(hh')$,
\end{enumerate}

We now show that for any $(\agent, \environment)$ pair and $t\in \mathbb{N}_0$, there will always be a pair $(h,h')$ that satisfies all four properties.

% First property.
\textit{(i) $hh' \in \realizablehistabbr_{t:\infty}$.}

Let us consider the first property. Recall that $\realizablehistabbr_{t:\infty}$ contains histories of length $t$ or greater that occur with non-zero probability from the interaction between $\agent$ and $\environment$. By definition of this set, it is non-empty. \checkmark

% Second and third properties.
\textit{(ii) $|h| \geq t$ and (iii) $|h'| > 0$.}

Now, observe that by necessity some of the $(h,h')$ pairs that satisfy the first condition, also satisfy the second and third conditions: Pick any realizable history from $\realizablehistabbr_{t:\infty}$ and there is a way to divide into pieces $h$ and $h'$ such that $|h| \geq t$ and $|h| > 0$. \checkmark

% Fourth conditions.
\textit{(iv) $\vec{\learnf}(h) = \vec{\learnf}(hh')$.}

Now, lastly, observe that some $(h,h')$ pairs that satisfy these three conditions must also ensure that the agent occupies the same agent state after experiencing $h$ and $hh'$. That is, that $\vec{\learnf}(h) = \vec{\learnf}(hh')$. If this were not the case, then for all such history pairs $h$ and $h'$ that satisfy conditions (i, ii, iii), the agent would have to occupy a \textit{different} agent state across $h$ and $hh'$. But then as the length of $h$ and $hh'$ go to infinity, the agent will occupy infinitely many different agent states, which contradicts its boundedness. \checkmark

This completes the argument for each of the four properties, and thus completes the proof. \qedhere

\end{dproof}

% Theorem: Properties of delta_t
%
\begin{customthm}{4.2}
For every $(\agent, \environment)$ pair:
\begin{enumerate}[(a)]
    \item The sequence $\{\delta_t(\agent, \environment)\}_{t=0}^\infty$ is non-increasing.
    
    \item $0 \leq \delta_t(\agent, \environment) \leq (\vmax - \vmin), \forall_{t \geq 0}$.
    
    \item $\delta_\infty(\agent, \environment)$ exists.
\end{enumerate}
\end{customthm}

% Proof.
\begin{dproof}[\cref{thm:delta_t_properties}]
We prove each property separately.

% (a) Sequence is non-increasing.
%
\textit{(a) The sequence $\{\delta_t(\agent, \environment)\}_{t=0}^\infty$ is non-increasing.}

Note that the set $\historiesastate_t$ induces a sequence of non-empty sets $\historiesastate_0$, $\historiesastate_1 \ldots$ such that
\begin{equation}
\label{proof_eq:realiz_h_subset}
\historiesastate_{t+1} \subseteq \historiesastate_{t},
\end{equation}
for all $t \geq 0$. Further, \cref{lem:hcirc_non_empty} ensures each of these sets are non-empty. The non-increasing nature of the sequence $\{\delta_t(\agent, \environment)\}_{t=0}^\infty$ holds by immediate consequence of this subset relation, as the $\sup$ ensures that
\[
\sup \mc{X}_{t+1} \leq \sup \mc{X}_{t} ,
\]
for countable sets $\mc{X}_{t+1}$ and $\mc{X}_t$ where $\mc{X}_{t+1} \subseteq \mc{X}_t$. Therefore, $\delta_{t+1}(\agent, \environment) \leq \delta_t(\agent,\environment)$ for all $t \geq 0$. \checkmark.

\vspace{5mm}

% (b) Bounds
%
\textit{(b) $0 \leq \delta_t(\agent, \environment) \leq (\vmax - \vmin),\ \forall_{t \geq 0}$.}

The lower bound holds as a direct consequence of the presence of the absolute value in \cref{eq:distortion_time_t}. The upper bound holds as a direct consequence of the boundedness of the function $\valuef$. \checkmark

\vspace{5mm}
% (c) Limit exists.
%
\textit{(c) The quantity $\delta_\infty(\agent, \environment)$ exists for every agent-environment pair.}

First we expand the definition of $\delta_\infty(\agent, \environment)$,
\begin{align}
&\delta_\infty(\agent, \environment)\\
&= \lim_{t \ra \infty} \delta_t(\agent, \environment)  \nonumber\\
&= \lim_{t \ra \infty} \sup_{(h,h') \in \historiesastate_t}
|v(\agent, \environment \mid h) - v(\agent, \environment \mid hh') |. \nonumber
\end{align}

Note that the sequence $\{\delta_t(\agent, \environment)\}_{t=0}^\infty$ is bounded by property (b): It is bounded below by 0 and above by $(\vmax - \vmin)$. Thus, since the $\limsup$ of a bounded sequence always exists, we conclude that $\delta_\infty(\agent, \environment)$ always exists. \checkmark

This completes the argument for all three statements, and thus completes the proof. \qedhere
\end{dproof}

% Proposition.
%
\begin{customprop}{4.3}
The following two statements hold:
\begin{enumerate}[(a)]

    % Bandits.
    %
    \item (Bandits) If an agent $\agent$ uniformly converges in a $k$-armed bandit $\environment$ in the sense that,
     \begin{equation}
        \exists_{t \in \mathbb{N}_0} \exists_{\rho \in \Delta(\actions)} \forall_{h \in \realizablehistabbr_{t:\infty}}\ \rho = \agent(h),
    \end{equation}
    then $\delta_\infty(\agent, \environment) = 0$.
    
    % MDPs.
    %
    \item (MDPs) If an agent $\agent$ uniformly converges in an MDP $\environment$ in the sense that,
    \begin{equation}
        \exists_{t \in \mathbb{N}_0} \exists_{\pi_\observations \in \Pi_\observations} \forall_{hao \in \realizablehistabbr_{t:\infty}}\ \pi_\observations(o) = \agent(hao),
        \label{eq:mdp_converge_policy2}
    \end{equation}
    then $\delta_\infty(\agent, \environment) = 0$.
\end{enumerate}
\end{customprop}

% Proof.
\begin{dproof}[\cref{prop:bandmdp_conv_implies_pconv}]

We prove the result for the MDP case, and since bandits are a clear instance of MDPs and the memoryless policies on $\mc{O} = \{o_1\}$ are equivalent to the set $\Delta(\actions)$, it follows that $\delta_\infty(\agent,\environment) = 0$ for bandits, too.

\textit{(b) (MDPs) If an agent $\agent$ converges in an MDP $\environment$ in the sense that,}
    \begin{equation}
        \exists_{t \in \mathbb{N}_0} \exists_{\pi_\observations \in \Pi_\observations} \forall_{hao \in \realizablehistabbr_{t:\infty}}\ \pi_\observations(o) = \agent(hao),
        \label{proof_eq:mdp_converge_policy}
    \end{equation}
\textit{then $\delta_\infty(\agent, \environment) = 0$.}

By \cref{proof_eq:mdp_converge_policy}, we know that there is a time $t$ at which the agent will act according to a fixed memoryless policy forever after.

Let us consider some time $t' >t$, and recall that the distortion at time $t'$ is given by,
%\begin{multline*}
\[
\delta_{t'}(\agent, \environment) = \sup_{(h,h') \in \historiesastate_{t'}} |v(\agent, \environment \mid h) - v(\agent, \environment \mid hh')| > 0.
\]
Let us consider what happens at the history pair, $h$ and $hh'$. Since $(h,h') \in \historiesastate_{t'}$, we know that the agent occupies the same agent state at both $h$ and $hh'$. Further, since $t' > t$, we know that the agent is equivalent to some memoryless policy at both $h$ and $hh'$, and will thus make the same choice of action distribution at both $h$ and $hh'$. Therefore, the agent will act in an equivalent manner at both $h$ and $hh'$. But this is also true for all subsequent realizable histories, and thus, the agent's performance \textit{must} be equivalent at both $h$ and $hh'$. We conclude that $v(\agent, \environment \mid h) = v(\agent, \environment \mid hh')$, and thus, for all $t' > t$,
\begin{equation}
    \delta_t(\agent, \environment) = 0.
\end{equation}
By the non-increasing nature of $\{\delta_t(\agent, \environment)\}_{t=0}^\infty$, we conclude $\delta_\infty(\agent, \environment) = 0$. \qedhere
\end{dproof}

% Proposition.
%
\begin{customprop}{4.4}
Any $\agent$ and MDP $e$ that satisfy
\begin{equation}
    \label{ap_eq:mdp_astate_delta_condition}
    \learnf(s,a,o) = \learnf(s',a',o')\ \Rightarrow\ o = o',\hspace{8mm} \forall_{s,s'\in\states}, \forall_{a,a'\in \actions}, \forall_{o,o'\in\observations},
\end{equation}
ensure $\delta_t(\agent, \environment) = 0$ for all $t \in \mathbb{N}_0$, and thus, $\delta_\infty(\agent, \environment) = 0$.
\end{customprop}

% Proof.
\begin{dproof}[\cref{prop:delta_is_0_mdp_astate}]
Consider a fixed but arbitrary pair $(\agent, \environment)$ such that \cref{ap_eq:mdp_astate_delta_condition} is satisfied.

Then, for arbitrary $t$, consider the performance of this agent at two histories $h$ and $hh'$, where $(h,h') \in \historiesastate_t$. Again we know such a pair exists by \cref{lem:hcirc_non_empty}. Further, we know by definition of $\historiesastate_t$ that the agent will occupy the same agent state after experiencing $h$ and $hh'$. That is, letting
\begin{align*}
    h &= \ldots ao,\hspace{8mm} s = \vec{\learnf}(h), \\
    h' &= \ldots a' o',\hspace{6mm} s' = \vec{\learnf}(h'),
\end{align*}
it follows that
\begin{equation}
    %\learnf(s_t, a_t, o_t) = u(s_{t'}, a_{t'}, o_{t'}),
    %\learnf(s_{t-1}, a_{t-1}, o_t) = u(s_{t-1'}, a_{t-1'}, o_{t'}).
    \vec{\learnf}(s, a, o) = \vec{\learnf}(s',a',o').
    \label{proof_eq:astate_same}
\end{equation}

By assumption, we supposed that the given agent and environment satisfy \cref{ap_eq:mdp_astate_delta_condition}, and therefore, together with \cref{proof_eq:astate_same}, it follows that $o = o'$. But, since $\environment$ is an MDP, anytime the MDP occupies the same MDP-state \textit{and} the agent occupies the same agent-state, the resulting performance will be the same. Thus, $\valuef(\agent, \environment \mid h) = \valuef(\agent, \environment \mid hh')$.

Since the time $t$ and history pair $(h,h')$ were chosen arbitrarily, we know that
\[
\valuef(\agent, \environment \mid h) = \valuef(\agent, \environment \mid hh'),
\]
holds for all $t$ and $h,h' \in \historiesastate_t$. Therefore, for all $t \in \N_0$
\begin{align*}
\delta_t(\agent,\environment) &= \sup_{(h,h') \in \historiesastate_t} 
    |\valuef(\agent, \environment \mid h) - \underbrace{\valuef(\agent, \environment \mid hh')}_{=\valuef(\agent, \environment \mid h)}|, \\
    &= \sup_{(h,h') \in \historiesastate_t} 
    |\valuef(\agent, \environment \mid h) - \valuef(\agent, \environment \mid h)|, \\
    &= 0,
\end{align*}
and hence, $\delta_\infty(\agent, \environment) = 0$. \qedhere
\end{dproof}

% Proposition.
%
\begin{customprop}{4.5}
There is a choice of MDP $\environment$ and bounded Q-learning $\agent$ that yields $\delta_\infty(\agent, \environment) > 0$.
\end{customprop}

\begin{dproof}[\cref{prop:pconv_nonstat_env}]

% Note: Proof strategy works for *many* envs.
%
For concreteness, we describe the agent and environment in more detail, but note that there is a neighborhood of such choices for which the same result holds. For simplicity, we let $\valuef$ capture the expected myopic reward (so discounted return with $\gamma=0$). The same result extends for other settings of $\gamma$, but the argument is simplest in the myopic case.

% MDP.
%
The MDP $\environment$ is as follows.
\begin{enumerate}[($e$.i)]
    \item The interface is defined as:
    \[
    \mc{O} = \{o_1, o_2\}, \hspace{6mm} \mc{A} = \{a_{\mr{move}}, a_{\mr{stay}}\}.
    \]
    Thus, since the environment $\environment$ is an MDP, we understand the MDP to have two states defined by the two observations $\{o_1, o_2\}$. We refer to these as MDP states throughout the proof.
    
    \item The MDP's transition function is defined as a deterministic function where $a_{\mr{move}}$ moves the agent to the other state, while $a_{\mr{stay}}$ causes the agent to stay in the same state.
    
    \item The MDP's reward function is as follows:
    \begin{equation}
        \label{proof_eq:nonstat_reward}
        r(oao') = \begin{cases}
        -1& o = o_1, \\
        +1& \text{otherwise}. \\
        \end{cases}
    \end{equation}
\end{enumerate}

% Agent.
%
The agent is as follows.
\begin{enumerate}[($\agent$.i)]
    \item The $\agent$ is an instance of tabular Q-Learning with $\varepsilon$-greedy exploration that experiences every $(o,a)$ pair infinitely often.
    
    \item Additionally, the agent $\agent$ is bounded in that it has finitely many agent states. While there is a vast space of ways to make such an agent bounded, we choose to consider the case where the agent state is comprised of three quantities, $s = (Q_t, \alpha_t, \varepsilon_t)$, where $Q_t : \observations \times \actions \ra [\vmin, \vmax]$ is the Q function, $\alpha_t \in [0,1]$ is the time-dependent step size, and $\varepsilon_t \in [0,1]$ is the time-dependent exploration parameter. Note that due to the boundedness of the agent, it can only have finitely many choices of $\alpha_t$ and $\epsilon_t$. Notice that as a result of this boundedness, it is an open question as to whether the precise conditions needed for Q-learning convergence in the classical sense will go through.

    \item Lastly, the agent violates \cref{ap_eq:mdp_astate_delta_condition}---The agent does not store its Q function as an exact function of the MDP's state, $o \in \mc{O}$, but rather, the agent's state-update function collapses $o_1$ and $o_2$ to a single Q function expressing the expected discounted return of $a_{\mr{move}}, a_{\mr{stay}}$.
\end{enumerate}

% Overview: two steps.
%
Now, we show that the pair defined above satisfies $\delta_\infty(\agent, \environment) > 0$. We do so through two points.  First, we show that for arbitrarily chosen $t$, there is a valid history pair $(h,h') \in \historiesastate_t$ where $h$ ends in $o_1$ and $h'$ ends in $o_2$, and (2) that the performance $\valuef(\agent, \environment \mid h) \neq \valuef(\agent, \environment \mid hh')$.

% (1) For all t, exists_{h,h' in historiesastate_t
%
\textit{(1) For any $t \in \mathbb{N}$, there is a valid $(h,h') \in \historiesastate_t$.}

Consider a fixed but arbitrary time $t \in \mathbb{N}$, and a history pair, $(h,h') \in \historiesastate_t$. Recall that this pair ensures that (1) $|h| \geq t$, (2) Both $h$ and $hh'$ occur with non-zero probability via the interaction of $\agent$ and $\environment$, and (3) The agent occupies the same agent state in $h$ and $hh'$. Further, we require that $h$ ends in $o_1$, and $h'$ ends in $o_2$. To see that such a history pair will exist for the chosen $t \in \mathbb{N}$, recall first that by assumption in point ($\agent$.i), Q-learning will experience every $(o,a)$ pair infinitely often. By the agent's boundedness in point ($\agent$.i) together with \cref{lem:hcirc_non_empty} ensures that $|\historiesastate_t| \geq 0$. Next, by point ($\agent.iii$), the agent is insensitive to the latest observation emitted, and so can occupy the same agent state between $h$ and $hh'$, even when $h$ ends in $o_1$ and $h'$ ends in $o_2$. For example, let $h = o_1a \ldots o_1ao_2$, and $h' = ao_2$. Note that $r(h) = r(o_1ao_1) = +1$, and note that $r(hh') = r(o_1ao_2) = +1$. Therefore the reward stream will not change the agent's internal state, and thus $\vec{\learnf}(h) = \vec{\learnf}(hh')$. \checkmark

% (2) perf(h) != perf(hh').
%
\textit{(2) $\valuef(\agent, \environment \mid h) \neq \valuef(\agent, \environment \mid hh')$.}

Next notice that for this chosen $(h,h')$ pair from step (1), the performance $\valuef(\agent, \environment \mid h) \neq \valuef(\agent, \environment \mid hh')$. To see why, observe that in history $h$, the agent occupies MDP state $o_1$, and thus, the subsequent reward is guaranteed to be $+1$ by \cref{proof_eq:nonstat_reward}. Conversely, in history $hh'$, the agent occupies MDP state $o_2$, and so the next reward receives is guaranteed to be $-1$ by \cref{proof_eq:nonstat_reward}. Since we assumed $\gamma=0$, it follows that $v(\agent, \environment \mid h) \neq v(\agent, \environment \mid hh')$ and consequently, $\delta_t(\agent, \environment) > 0$. \checkmark

Since $t$ was chosen arbitrarily, we note that the above property holds for \textit{all} $t \in \mathbb{N}$, and therefore,
\[
\lim_{t \ra \infty} \delta_t(\agent, \environment) > 0. \qedhere
\]
\end{dproof}

% Prop: Convergence time differs.
%
\begin{customprop}{4.6}
For any $(\agent, \environment)$ pair and any choice of $\beta, \epsilon \in \mathbb{R}_{\geq 0}$ let,
\begin{align*}
    % Behavior
    %
    t_\beta(\agent, \environment) &= \min \{t \in \mathbb{N}_0 : |c_t(\agent, \environment) - c_\infty(\agent, \environment)| \leq \beta\}, \\
    %
    % Performance
    t_\epsilon(\agent,\environment) &= \min \{t \in \mathbb{N}_0 : |\delta_t(\agent, \environment) - \delta_\infty(\agent, \environment)| \leq \epsilon\}.
\end{align*}
Then, for $\beta=0$, $\epsilon=0$, there exists pairs $(\agent, \environment)$ and $(\agent', \environment')$ such that
    \begin{equation}
        t_\epsilon(\agent,\environment) < t_\beta(\agent, \environment),
        \hspace{4mm}
        t_\beta(\agent',\environment') < t_\epsilon(\agent', \environment').
    \end{equation}
\end{customprop}

% % Proof.
\begin{dproof}[\cref{prop:conv_time_diff}]
We construct two examples $(\agent, \environment)$ where the time of performance convergence is different from the time of behavior convergence.

\textit{(a) Example 1: There exists a $(\agent, \environment)$ such that $t_\epsilon(\agent, \environment) < t_\beta(\agent, \environment).$} \\

We construct the example as follows. Let:
\begin{itemize}
    \item $\agent$ denote a single state agent that plays the same action at every round of interaction,
    
    \item $\environment$ denote an environment that produces $r(h_t) = 0$ for any history when $t > 10$, but $r(h_t) = 1$ for all histories where $t \geq 10$,
    
    \item And, the performance $\valuef(\agent, \environment \mid h)$ produces the reward received only in the next time step following $h$, for any $h$. 
\end{itemize}

Thus, since $\agent$ only has a single state, we know $c_0(\agent, \environment) = c_\infty(\agent, \environment) = 1$, and there $t_\epsilon=0$. 

Conversely, we can see that $\delta_0(\agent,\environment) = 1$ whereas $\delta_{10}(\agent, \environment) = \delta_\infty(\agent, \environment) = 0$, as the agent's performance is zero at $h_0$, but the agent's performance will be 1 for any $h_0 h'$ where $|h'| \geq 10$. Thus, since $t_\epsilon = 0$ and $t_\beta = 10$, we conclude $t_\epsilon < t_\beta$. \checkmark

\vspace{4mm}

\textit{(b) Example 2: There exists a $(\agent, \environment)$ such that $t_\beta(\agent, \environment) < t_\epsilon(\agent, \environment).$} \\

Consider an environment with a constant reward function, $r(h) = 0,\ \forall_{h \in \histories}$. Then, by immediate consequence, $\delta_0(\agent, \environment) = \delta_t(\agent, \environment) = \delta_\infty(\agent, \environment) = 0,\ \forall_{t \geq 0}$. Thus, $t_\epsilon =0$ for any agent in this environment.

Now we construct an agent for which the time of behaviour convergence $t_\beta(\agent, \environment) > 0$. Observe that this is true of an agent that switches between two agent states for the first ten time steps, choosing a different action in each state. Then, at time step $t=10$, the agent sticks with a single state and choice of action forever after. Consequently, we see that the agent requires two states to produce its behavior up until time $t=10$, at which time the agent only requires one state. Therefore, $c_0(\agent,\environment) = 2$, but $c_{10}(\agent, \environment) = c_\infty(\agent, \environment) = 1$. Hence, $t_\beta(\agent, \environment)=10$, whereas $t_\epsilon(\agent, \environment)=0$, and so $t_\epsilon < t_\beta$.  \checkmark

This completes the proof of both conditions, and we conclude. \qedhere
\end{dproof}

\end{document}